\documentclass[letterpaper, 10 pt, conference]{ieeeconf}  

\IEEEoverridecommandlockouts                              

\makeatletter
\let\NAT@parse\undefined
\makeatother
\usepackage[numbers,sort,comma,square]{natbib}
\usepackage{amsmath}
\usepackage{pgfplots}                      
\usepackage[pdftex,colorlinks=false,bookmarksopen=true,bookmarksopenlevel=0,pdfborder={0 0 0},breaklinks]{hyperref}
\usepackage[caption=false]{subfig}     
\usepackage[nolist]{acronym}           
\usepackage{bm}
\usepackage{ifmtarg}                   
\usepackage{siunitx}
\usepackage{graphicx}
\usepackage{multirow}                  
\usepackage{xstring}
\usepackage{stringstrings}

\begin{acronym}
  \acro{ADL}{Activities of Daily Living}
  \acro{CV}{cross validation}
  \acro{CNN}{Convolutional Neural Network}
  \acro{DWT}{Discrete Wavelet Transform}
  \acro{KRLS}{Kernel \acl{RLS}}
  \acro{MDWT}[mDWT]{marginal \acl{DWT}}
  \acro{MER}{Movement Error Rate}
  \acro{MEAN}{mean value}
  \acro{NINAPRO}[NinaPro]{Non-Invasive Adaptive Prosthetics}
  \acro{RBF}{Radial Basis Function}
  \acro{RLS}{Reguralized Least Squares}
  \acro{RMS}{Root Mean Square}
  \acro{SEMG}[sEMG]{surface electromyography}
\end{acronym}

\makeatletter
\newcommand{\isempty}[3]{\@ifmtarg{#1}{#2}{#3}}
\newcommand{\isnotempty}[2]{\@ifnotmtarg{#1}{{#2}}}
\makeatother

\newcommand{\func}[2]{#1\isnotempty{#2}{\left(#2\right)}}    

\renewcommand{\vec}[1]{\bm{#1}}  
\newcommand{\mat}[1]{\bm{#1}}

\newcommand{\lnorm}[1]{\left\lVert#1\right\rVert}

\title{\LARGE \bf
Visual Cues to Improve Myoelectric Control of Upper Limb Prostheses}

\author{Andrea~Gigli$^{1}$, Arjan~Gijsberts$^{1}$, Valentina~Gregori$^{1}$, Matteo~Cognolato$^{2}$, Manfredo~Atzori$^{2}$, and Barbara~Caputo$^{1}$
\thanks{*This work was supported by the Swiss National Science Foundation Sinergia project \#160837 ``Megane Pro''}
\thanks{$^{1}$Department of Computer, Control, and Management Engineering, University of Rome La Sapienza, via Ariosto 25, 00185 Roma, Italy 
        {\tt\small surname@dis.uniroma1.it}}%
\thanks{$^{2}$Department of Business Information Systems, University of Applied Sciences Western Switzerland (HES-SO Valais), Sierre, Switzerland {\tt\small name.surname@hevs.ch}}%
}

\begin{document}

\maketitle
\thispagestyle{empty}
\pagestyle{empty}

\begin{abstract}
The instability of myoelectric signals over time complicates their use to control highly articulated prostheses.
To address this problem, studies have tried to combine surface electromyography with modalities that are less affected by the amputation and environment, such as accelerometry or gaze information.
In the latter case, the hypothesis is that a subject looks at the object he or she intends to manipulate and that knowing this object's affordances allows to constrain the set of possible grasps.
In this paper, we develop an automated way to detect stable fixations and show that gaze information is indeed helpful in predicting hand movements. 
In our multimodal approach, we automatically detect stable gazes and segment an object of interest around the subject's fixation in the visual frame.
The patch extracted around this object is subsequently fed through an off-the-shelf deep convolutional neural network to obtain a high level feature representation, which is then combined with traditional surface electromyography in the classification stage.
Tests have been performed on a dataset acquired from five intact subjects who performed ten types of grasps on various objects as well as in a functional setting.
They show that the addition of gaze information increases the classification accuracy considerably.
Further analysis demonstrates that this improvement is consistent for all grasps and concentrated during the movement onset and offset.

\end{abstract}

\section{Introduction} \label{sec:introduction}

The loss of a hand or an arm due to amputation has a drastic impact on the quality of life.
Although advanced myoelectric prostheses have the potential to restore some of the lost functionality, their acceptance among amputees is very low~\cite{peerdeman11}.
Aside from high cost, one of the problems with these active prostheses is that their control is not robust and requires a long and painful training procedure.
Myoelectric signals change over time, for instance due to electrode shift, user adaptation, and fatigue, and this hurts control robustness.  

Academic efforts have therefore started to focus on how to make prosthetic control more stable and more intuitive.
An interesting avenue is to reduce the dependency on \ac{SEMG} by including other sources of contextual information, such as inertial sensors~\cite{fougner11} or computer vision~\cite{dosen10,dosen11}.
The working principle is that this context is helpful in decoding the intent of the prosthesis user.
This seems obvious in case of the orientation of the limb (cf. inertial sensors), but also the user's gaze behavior and the visual description of an object of interest may contain important side-information to determine the desired hand movement.
For example, it seems more likely that a person that is fixating a pen lying on a table desires to perform a writing tripod than a power disk grasp.

Recent studies have investigated the use of visual information to preshape a prosthesis based on the estimated object size and orientation.
Rather than object size and orientation, we argue that also the object's affordances are relevant to determine the desired grasp type.
We therefore extract high-level features of the object of interest using a powerful, off-the-shelf convolution neural network.
These features are highly discriminative for object identification and they will therefore also contain informative content on the object's functionality. 
Furthermore, in contrast to earlier studies we do not require users to trigger the visual recognition system, but instead use gaze tracking to automatically detect stable fixations and to segment the object of interest. 

The proposed method was evaluated offline on data collected from five intact subjects performing ten grasps.
All grasps were repeated both seated and standing, and with three different objects each.
The chosen objects are associated with activities of daily living and thus representative of a home environment.
To promote variability in the arm dynamics and visual scene, subjects also performed these grasps as part of 15 functional movements (e.g., open a zipper using a lateral grasp).


The remainder of this paper continues with an overview of related work in \autoref{sec:related}. In \autoref{sec:method}, we give a detailed description of our method to automatically detect fixations and how to integrate the object's visual representation with \ac{SEMG}. We then describe the experimental setup of our evaluation in \autoref{sec:setup} and follow this with the results in \autoref{sec:results}. This paper is concluded in \autoref{sec:conclusions}.  

\section{Related Work} \label{sec:related}

The difficulty of reliably measuring and interpreting \ac{SEMG} has led to active research on the inclusion of other types of sensory modalities to control myoelectric prostheses, such as sonomyography, mechanomyography, and force myography~(for detailed overviews, see~\cite{fang15,loboprat14}). 
Besides those that measure muscular activity, also modalities that provide an informative context on the intended movement have been combined with \ac{SEMG}.
Several studies have shown that accelerometry of the relevant arm provides useful information on arm orientation and dynamics that is complementary to \ac{SEMG}~\cite{fougner11,gijsberts13}.

More recently, also computer vision and gaze information have been considered to improve intent recognition.
Their relevance has been shown in early studies, in which innovative systems were proposed for controlling the prehension of a transradial prosthesis.
These either used a webcam~\cite{dosen10,dosen11} or electro-oculography~\cite{hao13} to automatically preshape the prosthesis based on the estimated object size and orientation.
This approach has subsequently been integrated with a myoelectric control strategy by \citet{markovic14,markovic15}.
In their system, myoelectric activity is combined with computer vision and inertial sensing to provide artificial proprioceptive feedback on the grasp type and object size.
Via sEMG-based sequential and proportional control, the user can override the automated preselections of the system.
The use of computer vision in the context of prosthetics was also hinted at by \citet{ghazaei15}, who used deep learning to classify grasps based on the object's appearance.

Slightly different from our application in prosthetics for amputees, \ac{SEMG} and gaze information has also been used to operate a robot arm for tetraplegic patients. \citet{corbett14} use the subject's gaze to help to determine the target position of a reaching movement, while \citet{mcmullen14} combine this with computer vision to initiate and automatically perform the reach-grasp-drop motion of the robot arm.

\section{Gaze Integration} \label{sec:method}


The basic idea behind this work is to extract a representation of the object that is observed during a prehension and use it as an auxiliary cue in support of a standard \ac{SEMG} based grasp classifier.
To do so, we designed a method to automatically detect stable gaze fixations, extract relevant visual information associated with those fixations, and subsequently integrate this information in the movement classifier.

\subsection{Fixation detection} \label{sec:method:detection}
\begin{figure}
\centering
\includegraphics[width=0.5\textwidth]{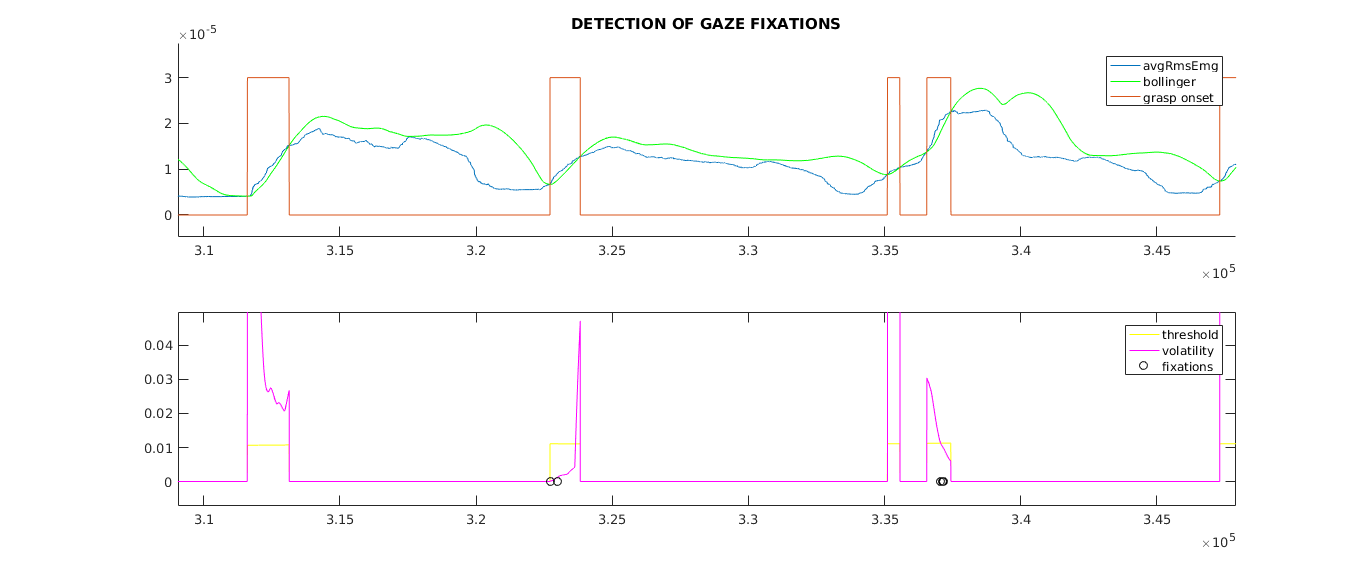}
\caption{Example of gaze fixation detection. The grasps' onsets (red) are identified in the \textit{AvgRmsEmg} signal (blue) by thresholding it with its upper Bollinger Band (green). During the onsets, fixations (black circles) are identified when the gaze volatility (magenta) falls below a certain threshold (yellow). The figure is best viewed in color.} \label{fig:method_summary}
\end{figure}

The first step of the algorithm consists in finding fixations in the gaze tracking data. A fixation consists of a period of time (generally between \SIrange{350}{450}{\milli\second}~\cite{johansson01}) where the eye-gaze remains in a limited area of the visual field.
Since we are only interested in fixations that precede a grasp, we attempt to identify an increase in muscle activity by looking at the \ac{RMS} of the myoelectric signals in a sliding window of length $\tau_\textit{rms}$. 
As can be seen in \autoref{fig:method_summary}, the average activity over all electrodes, denoted with \textit{AvgRmsEmg}, increases drastically during the initial reach-to-grasp phase. We identify these increases in an online manner using Bollinger bands, which calculate the number of standard deviations that a current value $x_t$ of a signal is from a historical mean within a sliding window of length $\tau_\textit{boll}$
\begin{align}
  \func{b}{\vec{x}, t, \tau_\textit{boll}} & = \frac{x_t - \func{\mu}{\vec{x}_{t:t-\tau_\textit{boll}}}}{\func{\sigma}{\vec{x}_{t:t-\tau_\textit{boll}}}}\enspace,
\end{align}
where $\vec{x}_{t:t-\tau_\textit{boll}}$ denotes the sliding window, and $\func{\mu}{\cdot}$ and $\func{\sigma}{\cdot}$ denote the window mean and standard deviation. Since we are interested in sudden \emph{increases} in muscle activity, we limit our attention to when the value exceeds the upper Bollinger band
\begin{align}
  \func{b}{\vec{x}, t, \tau_\textit{boll}} & \geq \eta,
\end{align}
where $\eta$ regulates the sensitivity of the method to the signal's variations. \autoref{fig:method_summary} shows the identified time intervals associated to the grasp's onset (red) when \textit{AvgRmsEmg} exceeds its upper Bollinger band (green).

After identifying these regions where the hand has started reaching, we identify stable fixations on the basis of the gaze volatility. Since the gaze is represented as a 2-dimensional vector (both $x$ and $y$ coordinates in the image frame), we define multidimensional volatility of a sequence of gaze points $\mat{X}$ as Euclidean variance around the centroid 
\begin{align}
  \func{v}{\mat{X}, t, \tau_\textit{gaze}} = \frac{1}{\tau_\textit{gaze}} \sum_{i=t}^{t-\tau_\textit{gaze}} \lnorm{\vec{x}_i - \func{\vec{\mu}}{\mat{X}_{t:t-\tau_\textit{gaze}}}}_2 \enspace .
\end{align}
We use this quantity to define a fixation when volatility $\func{v}{\mat{X}, t, \tau_\textit{gaze}}$ falls below a threshold which is updated to the $40^\text{th}$ percentile of the volatility every \SI{0.5}{\second}. \autoref{fig:method_summary} shows the gaze volatility (magenta) along with its threshold (yellow) and the selected gaze fixations (black circles). 
Based on results during preliminary analyses, we have set the values of the parameters to $\tau_{\textit{rms}} = \tau_{\textit{boll}} = \tau_\textit{gaze} = \SI{300}{\milli\second}$ and $\eta=2$.

\subsection{Visual Feature Extraction} \label{sec:method:features}

For each fixation, the gaze position will be used to obtain an image of the observed object from the first-person video recording of the scene. From this image, the object's affordances will be encoded into appropriate visual features. 

The video recording and the gaze tracking data are synchronized and expressed in the same reference system, thus the gaze point always lies on the object on which the user is focusing. We isolate this object from the others in the image using the Active Segmentation algorithm by \citet{mishra12}. This method uses brightness, colors, and textures to segment the object on which the gaze falls. The drawback of the fixation-guided segmentation is its sensitivity to noise in the gaze position estimate. On the other end, its substantial advantage over object-detection methods based on machine learning is that it does not require any prior knowledge about the appearance of the objects of interest.

The object's affordances are extracted from its image and encoded into appropriate visual features using a \ac{CNN} as a feature extractor. Deep visual features, indeed, are able to gather spatial and high-level visual characteristic, like shapes and color gradients. The object image is fed into the VGG-16 \ac{CNN} pre-trained on ImageNet~\cite{simonyan14} and the activation of the second-last fully-connected layer is taken as the image visual feature. 

Under the assumption that the object of interest remains the same during the whole prehension, the \ac{CNN} feature associated with a certain fixation is maintained for all the subsequent samples, until the next fixation. 
A side effect of this choice is that each arm rest will be associated with the visual features of the object grasped in the previous prehension. However, this is inevitable because we do not know a priori the grasp's duration.

\subsection{Multimodal integration} \label{sec:method:integration}

At the end of the feature extraction process, the visual cue can be used alone or in conjunction with the myoelectric one to train a grasp classifier.
Among the possible methods to integrate multimodal cues, we opt for mid-level integration~\cite{tommasi08}, also known as integration at the kernel level.
This method combines couples $\left( \mathbf{x}, \mathbf{y} \right)$ of multimodal samples of the type $\mathbf{x} = \{ \mathbf{x}^1 \text{,} \cdots \text{,} \mathbf{x}^{C} \}$ by computing their similarity via a weighted sum of cue-specific kernel functions
\begin{align}
  \func{k_\textit{mc}}{\mathbf{x},\mathbf{y}} & = \sum_{i=1}^C{w_i k_i(\mathbf{x}^i, \mathbf{y}^i)} && \text{for } w_i>0\enspace .
\end{align}
The weights $w_i$ of the kernel combination are free hyperparameters of the multi-cue kernel.
Such similarities will be used by a kernel machine classifier to create the prediction model.



\section{Experimental Setup} \label{sec:setup}

We collected a custom dataset in which we recorded \ac{SEMG} and gaze while subjects performed a set of grasps on different objects. In the following, we detail the dataset and how the data was used in our offline evalutation. 

\subsection{Dataset} \label{sec:setup:dataset}

Five intact subjects (4M, 1F) participated in our study.
We selected ten grasps based on relevant literature~\cite{feix16} and on their perceived importance for \ac{ADL}.
Each of the grasps was performed on three representative objects that could reasonably be manipulated using the respective grasp.
In selecting these objects, we attempted to re-use them as much as possible for multiple grasps to enforce a many-to-many relationship: grasps can be used with multiple objects and objects can be used with multiple grasps.
This avoids the risk that an object's identity alone is sufficient to unequivocally predict a grasp.
During the acquisition, we made sure that there were always a minimum of five objects placed in front of the subject, to encourage realistic gaze behavior and to increase visual clutter. 

Aside from multiple objects, the acquisition protocol was extended in two other manners to encourage variability in the myoelectric signals.
One source of variability is given by the limb position effect, meaning that the signals will depend on the orientation of the limb.
We took this into account by performing all movements both while seated and standing, which are likely the most common orientations in \ac{ADL}.
Second, we extended the protocol with either one or two functional tasks for each of the grasps.
This introduces variability in the dynamic context of the hand, or more precisely crosstalk due to the added activity of muscles controlling the wrist and limb. 
Also in this case these functional movements were selected to represent \ac{ADL}.
The grasps, their respective objects, and the functional tasks are listed in \autoref{tab:setup:movements}.


\newcommand{\replaceunderscore}[1]{\StrSubstitute{#1}{_}{ }[\temp]\capitalizewords{\temp}}
\newcommand{\cincludegraphics}[2][]{\raisebox{-0.17\height}{\includegraphics[#1]{#2}}}
\newcommand{\multicolumntable}[9]{%
\multicolumn{1}{ |l  }{\multirow{3}{*} {\replaceunderscore{#1}} } &
\multicolumn{1}{ |c| }{\isnotempty{#2}{{\cincludegraphics[height=0.02\paperheight]{figures/movements/static__#1__#2.png}}}} & Take the \replaceunderscore{#2} while Seated/Standing & \isnotempty{#5}{{\cincludegraphics[height=0.02\paperheight]{figures/movements/functional__#1__#5.png}}} & #7 while #9 \\ \cline{2-5}
\multicolumn{1}{ |c  }{}                        &
\multicolumn{1}{ |c| }{\isnotempty{#3}{{\cincludegraphics[height=0.02\paperheight]{figures/movements/static__#1__#3.png}}}} & Take the \replaceunderscore{#3} while Seated/Standing & \isnotempty{#6}{{\cincludegraphics[height=0.02\paperheight]{figures/movements/functional__#1__#6.png}}} & \isnotempty{#6}{#8 while #9} \\ \cline{2-5}
\multicolumn{1}{ |c  }{}                        &
\multicolumn{1}{ |c| }{\isnotempty{#4}{{\cincludegraphics[height=0.02\paperheight]{figures/movements/static__#1__#4.png}}}} & Take the \replaceunderscore{#4} while Seated/Standing &  &  \\ \cline{1-5}
}

\begin{table*}[ptb]
\centering
\caption{Combination of grasps, objects, and functional tasks.}
\label{tab:setup:movements}
g\begin{tabular}[t]{ c | c | c | c | c |}
\cline{2-5} & \multicolumn{2}{ c| }{Static} & \multicolumn{2}{ c| }{Functional} \\[0.3cm] \cline{2-5}
\cline{1-5} \multicolumn{1}{ |c| }{Grasps} & Objects & Task Description & Objects & Task Description \\[0.3cm] \cline{1-5}
\multicolumntable{medium_wrap}{bottle}{can}{door_handle}{can}{door_handle}{Drink from the Can}{Open and close the Door Handle}{Standing}
\multicolumntable{lateral}{cup}{key}{pencil_case}{key}{jacket}{Turn the Key in the lock}{Open and close the Jacket}{Standing}
\multicolumntable{parallel_extension}{plate}{book}{drawer}{plate}{}{Lift the Plate}{}{Standing}
\multicolumntable{tripod_grasp}{bottle}{cup}{drawer}{bottle}{drawer}{Open and close the cap of the Bottle}{Open and close the Drawer}{Standing}
\multicolumntable{power_sphere}{ball}{light_bulb}{key}{ball}{}{Move the Ball to the right and back}{}{Standing}
\multicolumntable{precision_disk}{jam_jar}{light_bulb}{ball}{jam_jar}{light_bulb}{Open and close the lid of Jam Jar}{Screw and unscrew the Light Bulb}{Seated}
\multicolumntable{prismatic_pinch}{clothespin}{key}{can}{clothespin}{}{Squeeze the Clothespin}{}{Seated}
\multicolumntable{index_finger_extension}{remote_control}{knife}{fork}{remote_control}{knife}{Press a button on the Remote Control}{Cut bread with the Knife}{Seated}
\multicolumntable{adducted_thumb}{screwdriver}{remote_control}{wrench}{screwdriver}{}{Turn the Screwdriver}{}{Seated}
\multicolumntable{prismatic_four_fingers}{knife}{fork}{wrench}{fork}{}{Move the Fork to the right and back}{}{Seated}
\end{tabular}
\end{table*}

During the exercise, the subject was in front of a table on which the objects were placed.
Prior to each grasp, a screen showed short movies of the movements with the aim to clarify how each of the objects should be approached.
The scope of this video was to help the subject become familiar with the procedure and to perform a training trial while the video was playing.
After this initial phase, the subject was requested to repeat each movement-object combination or functional movement four times. The computer indicated when to start the grasp and when to release via audio instructions. 
As visual support, the required grasp was schematically shown on the screen for the entire duration of the exercise. 
Each repetition took approximately \SI{8}{\second}, containing the actual grasp (\SIrange{4}{5}{\second}) and the subsequent transition back to the rest posture (\SIrange{3}{4}{\second}).

\begin{figure}[tbp]
  \includegraphics[width=0.99\linewidth]{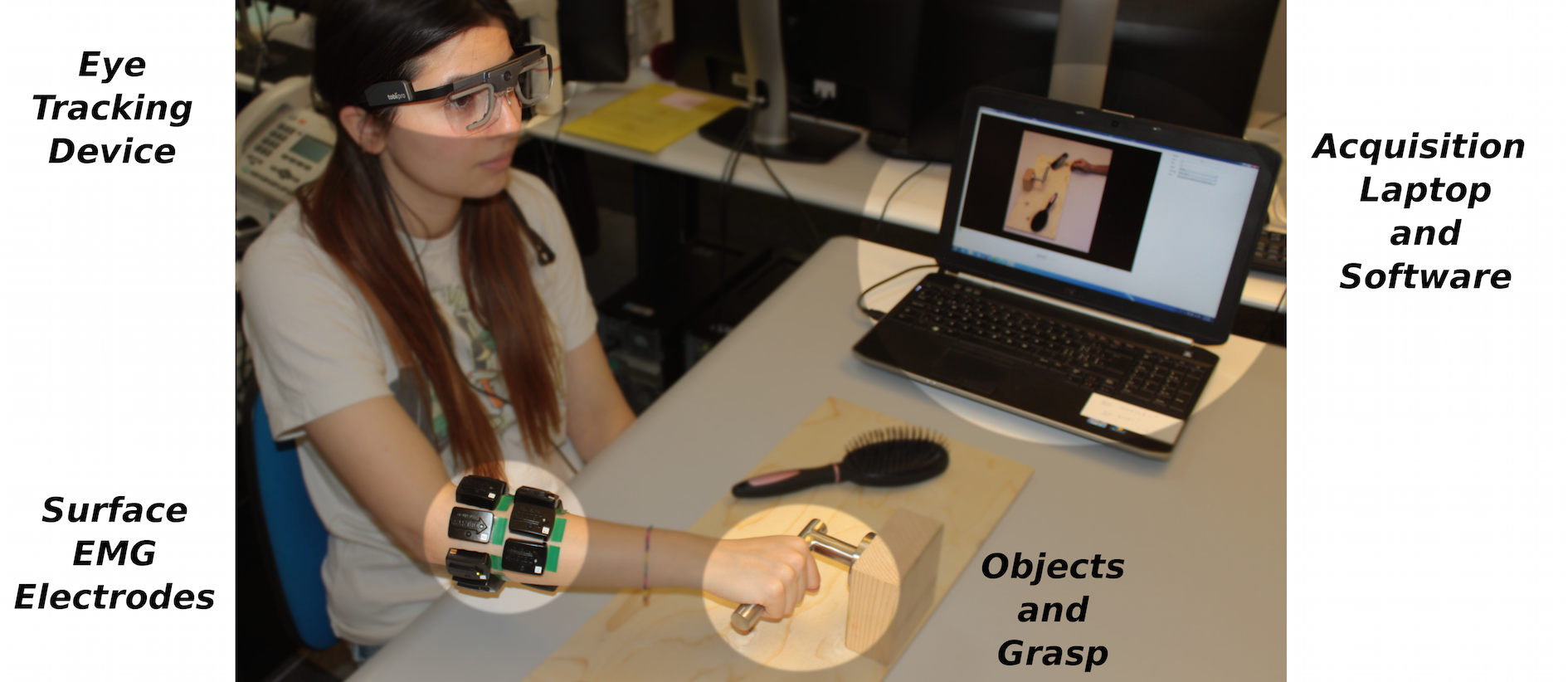}
  \caption{Overview of the acquisition setup, including the \ac{SEMG} electrodes, the gaze tracking device, and the laptop used for the stimulus.} \label{fig:setup:acquisition}
\end{figure}

Muscular activity was recorded using twelve Delsys Trigno double differential \ac{SEMG} electrodes placed in two rows around the forearm, where the upmost row contained eight electrodes and the remaining four were placed lower (see \autoref{fig:setup:acquisition}).
The myoelectric signals were sampled at \SI{2}{\kilo\hertz}.
At the same time, the gaze and first-person scene video were recorded using the Tobii Pro Glasses II. 
These glasses record the subject's gaze at \SI{100}{\hertz} with a theoretical accuracy and precision of \SI{0.5}{\degree} and \SI{0.3}{\degree} degrees \ac{RMS}, respectively.
The frame also contains a forward facing scene camera with a field of view of \SI{90}{\degree} that records Full HD video at \SI{25}{Hz}.
The onboard software of the Tobii glasses conveniently precomputes the gaze point in the reference frame of the gaze camera, which is what we will use in the remainder of the paper.
\autoref{fig:setup:acquisition} gives an overview of the acquisition setup.

The acquisition laptop was used to assign timestamps to \ac{SEMG} and gaze samples in a shared reference time.
These timestamps were used during preprocessing to synchronize all modalities and to upsample them to the sampling rate of \ac{SEMG}.
Furthermore, we filtered powerline interference and corrected the labels using the relabeling method described by \citet{gijsberts14tnsre}.

\subsection{Classifier} \label{sec:setup:classifier}

Also our classification setup was inspired by~\cite{gijsberts14tnsre}, based on a \ac{KRLS} classifier~\cite{rifkin03}.
This learning method is a so-called kernel method, meaning that it approaches nonlinear problems by using kernel functions that implicitly map the original input space into a high-dimensional feature space.
This also means that it is straightforward to use the multicue kernel described in \autoref{sec:method} in this classifier.   



Based on reports in previous work~\cite{gijsberts14tnsre}, we opted to combine the \ac{MDWT} representation for \ac{SEMG} in a sliding window of \SI{200}{\milli\second} with the exp-$\chi^2$ kernel function
\begin{align}
k_{\chi^2}(\mathbf{x},\mathbf{y}) & = \exp\left( -\gamma_{\chi^2} \sum_{i=1}^n{\frac{(x_i-y_i)^2}{x_i+y_i}} \right) \text{  for  } \gamma_{\chi^2}>0\enspace .
\end{align}

For the visual cue, we chose the standard \ac{RBF} kernel function
$$ k_{\textit{rbf}}(\mathbf{x},\mathbf{y}) = \exp\left( -\gamma_{\textit{rbf}} \lVert \mathbf{x}-\mathbf{y} \rVert^2 \right) \text{  for  } \gamma_{\textit{rbf}}>0\enspace .$$
A linear kernel is typically sufficient for the representation at high levels of a \ac{CNN}, but we prefer an \ac{RBF} kernel to ensure that the outputs of both kernels in the combination are in the range $[0, 1]$. The multi-cue kernel combining the myoelectric and the visual cues becomes therefore
$$k_{\textit{mc}}(\mathbf{x},\mathbf{y})=w_{\textit{emg}} k_{\chi^2}(\mathbf{x},\mathbf{y}) + w_{\textit{cnn}} k_{\textit{rbf}}(\mathbf{x},\mathbf{y}) .$$

The \ac{KRLS} algorithm and the multi-cue kernel require the optimization of the regularization parameter $\lambda$, the kernel-specific parameters $\gamma_{\chi^2}$ and $\gamma_{\textit{rbf}}$, and the weights used in kernel combination $w_{\textit{emg}}$ and $w_{\textit{cnn}}$. 
The parameters are optimized using k-fold cross-validation on the training set, where each of the folds corresponds to one of the movement repetitions used for training. The parameter ranges that we considered with a dense grid search are  
$\lambda \in \{ 2^{-14}, 2^{-12}, \cdots, 2^{-4} \}$, 
$\gamma_{\chi^2} \in \{ 2^{-14}, 2^{-12}, \cdots, 2^{-8} \}$, 
$\gamma_{\textit{rbf}} \in \{ 2^{-20}, 2^{-18}, \cdots, 2^{-14} \}$, and
$w_{\textit{emg}}, w_{\textit{cnn}} \in \{ 0, 0.1, 0.2, \cdots, 1 \}$ such that $w_{\textit{emg}}+w_{\textit{cnn}}=1$.
The grids have been determined during preliminary analyses.

The grasp classification is repeated over four possible training/test splits of the database, such that each of the four repetitions of a movement is used once to test the model while the remaining three are used as training set.
Subsequently, the prediction accuracy is averaged over the four splits.
For computational reasons, we subsampled the test data at a factor $20$, meaning that effectively we predict a sliding window with interval of \SI{10}{\milli\second}.
The training data instead was subsampled with an additional factor of $10$, while the data used for hyperparameter optimization was subsampled with a factor of $10 \cdot 4$.
Besides our multimodal classifier, we also include single cue classifiers as reference and a baseline that predicts simply the most common class in the training data.
In our specific case, this means predicting always the ``rest'' class, since this is the most common class due to our acquisition protocol. 

\section{Results} \label{sec:results}

\begin{figure}
\centering
\includegraphics[width=0.5\textwidth]{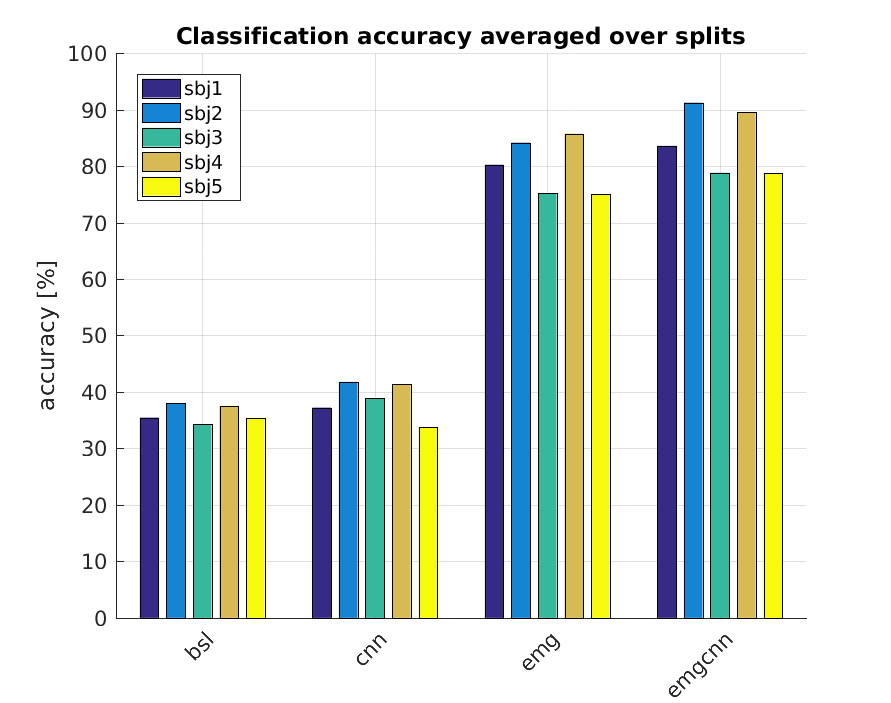}
\caption{Per-subject grasp classification accuracy for the Baseline, the CNN, the EMG and the EMG+CNN classifiers. All the accuracies are averaged over the training/test splits.} \label{fig:accuracy}
\end{figure}

The goal of this section is to determine if the standard \ac{SEMG} approach would benefit from the integration of the visual cues found by our algorithm. 
\autoref{fig:accuracy} reports the average classification accuracy of the four classifiers for each subject.
The sole visual cue does not produce a considerable improvement in accuracy with respect to the baseline, as the average improvement is of the 2\% and it is mainly due to two of the five subjects. 
Nevertheless, the integration of vision to the muscular cue increases the average accuracy of more than 4\% over that of the EMG classifier, and this appears to be a common trend for all the subjects. This result confirms our initial guess that the visual cue conveys complementary information with respect to the muscular one and that their integration improves the performance of the grasp classification task.

\begin{figure}
\centering
\includegraphics[width=0.5\textwidth]{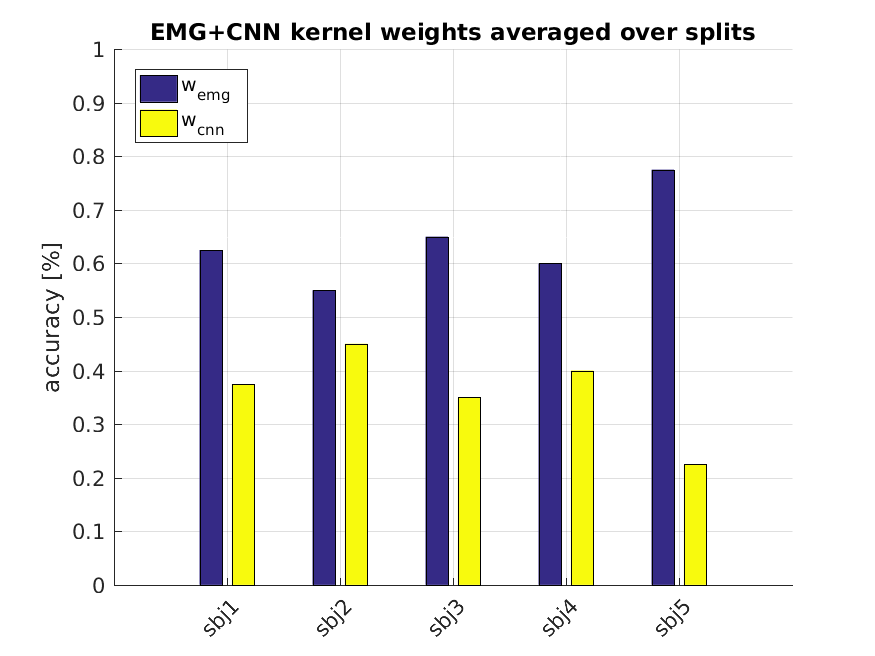}
\caption{Optimal kernel weights used to integrate the muscular and the visual cue. The weights are averaged over the training/test splits. Each kernel weight represents the contribution of the respective cue to the multi-cue classification.} \label{fig:weights}
\end{figure}

The contribution of each of the two cues to the multimodal classification is indicated by the values of the weights $w_{\textit{EMG}}$ and $w_{\textit{CNN}}$ that the algorithm automatically choose (during hyperparameter optimization) to combine the cues at the kernel level. \autoref{fig:weights} reports the values of the kernel weights for each subject and demonstrates how the contribution of the two cues is balanced, being around the 65\% for the muscular cue and the 35\% for the visual one on average.

\begin{figure}
\centering
\includegraphics[width=0.5\textwidth]{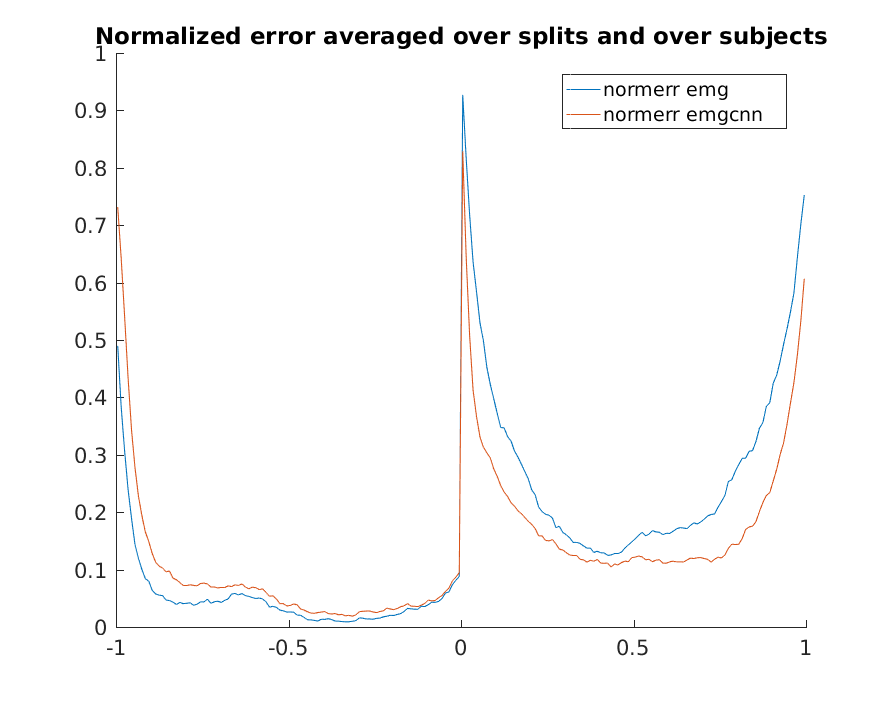}
\caption{Normalized Error of the EMG and the EMG+CNN classifiers averaged over the training/test splits and the subjects. The integration of vision to the EMG reduces the prediction error during the grasp, particularly during the onset and the offset of the movement, but slightly deteriorates the recognition of the rest class.} \label{fig:normerr}
\end{figure}

We also analyze the distribution of the prediction error during the different phases of the prehension. Each prehension of the experiment has a different duration after relabeling, but always consists of a grasp preceded by a rest phase. In \autoref{fig:normerr}, we report the prediction error with respect to the normalized duration of the rest phase ($[-1,0]$) and subsequent grasp ($[0,+1]$). 
The addition of visual cues consistently reduces the prediction error during the grasp ($t \in [0,1]$). Relevantly, the most consistent reduction in prediction error due to the visual cue (around the 10\%) happens at the onset and at the offset of the grasp. This indicates that vision compensates for the increased level of noise in the myoelectric signals during movement transitions.
At the same time, the visual cue causes a slightly higher prediction error during the rest phase. This is because, generally, the visual information related to arm rest comes from the previous grasp and is, in fact, misleading for the classification of the ``rest'' movement. As already explained in \autoref{sec:method:features}, this side-effect of the visual feature propagation is inevitable because we do not know in advance the duration of the grasps.

\begin{figure}
\centering
\includegraphics[width=0.5\textwidth]{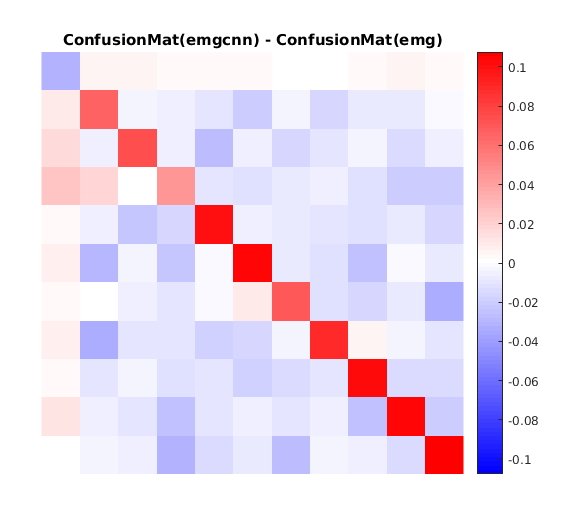}
\caption{ Difference between the confusion matrices of the EMG+CNN and the EMG classifier. Positive values on the diagonal indicate better recognition of the relative classes when integrating CNN to EMG. The figure is best viewed in color.} \label{fig:confmat}
\end{figure}

Qualitatively, the classification improvement obtained by integrating CNN and EMG can be observed by subtracting the confusion matrix of EMG+CNN to that of EMG. This difference is shown in \autoref{fig:confmat}. The positive values on the diagonal indicate a uniform improvement of the classification accuracy for all the 10 grasps. However, the negative value at location (1,1) indicates an increase in rest misclassification and confirms the considerations made about the effect of holding the visual cues also during rest. 

\begin{figure}
\centering
\includegraphics[width=0.5\textwidth]{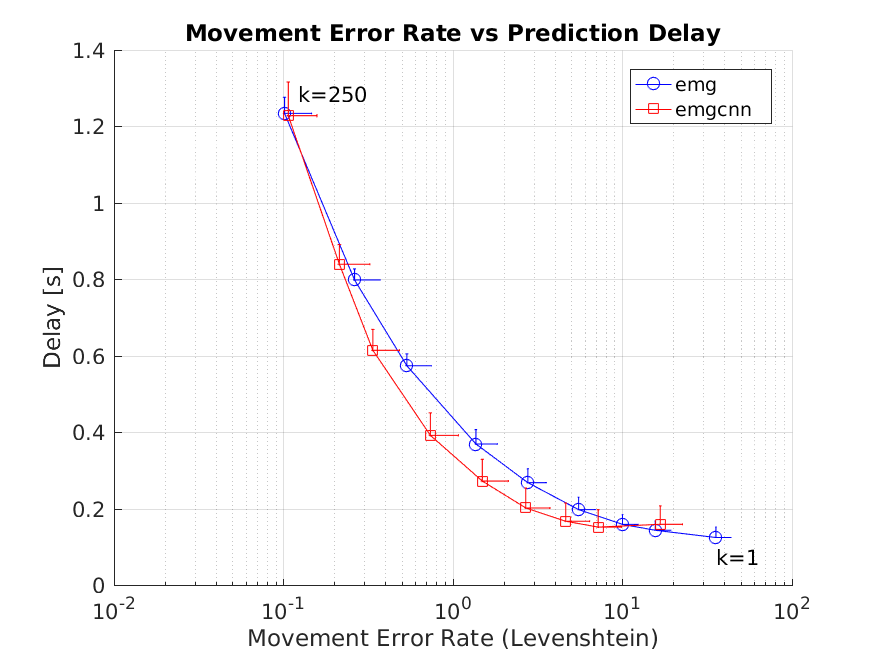}
\caption{Results of the EMG and the EMG+CNN classifiers in terms of \acl{MER} and prediction delay while varying the length $k$ of a majority vote window. The error bars indicate unit standard deviation.} \label{fig:merpd}
\end{figure}

In tasks where the classification predictions form a temporal sequence, it is advisable to define performance metrics that distinguish if errors are caused by misclassifications or delays in the predictions.
This is useful, for instance, to observe the effects of smoothing the series of predicted movements via a majority vote. When the dimension $k$ of the majority vote window is increased, the number of misclassifications generally decreases at the expense of a higher prediction delay. 
Standard classification accuracy fails to catch these competing effects, hence we will analyze the classifier performances also using the \ac{MER} and the prediction delay, proposed by \citet{gijsberts14tnsre}.
The \ac{MER} measures the similarity between the true and the predicted series of movements rather than considering the accuracy of the classification sample by sample. This quantity is insensitive to delays in the prediction, which are instead measured via the prediction delay, defined as the average time interval between a label change and the first correct prediction.
\autoref{fig:merpd} represents the values of \ac{MER} and delay achieved by the EMG (blue) and the EMG+CNN (red) classifiers when varying the length $k$ of a majority vote window ($k \in \{1, 3, 5, 11, 25, 50, 100, 150, 250\}$). The integration of visual features to muscular ones proves to reduce the \ac{MER} consistently for all the considered values of $k$. In particular, for $k<50$, EMG+CNN shows a halved \ac{MER} with an unchanged prediction delay. This shows that the reduction in error of the EMG+CNN classifier does not come at the cost of increased delay. 


\section{Conclusions} \label{sec:conclusions}

This work demonstrated how standard \ac{SEMG} based grasp classification benefits from the integration of the affordances of the manipulated objects. 
We proposed a method to automatically extract the object affordances from a first-person video recording of the scene and an estimate of the gaze position. 
The method identifies relevant gaze fixations on the base of ocular and muscular activity. The objects observed during such fixations are segmented and their affordances are encoded into high-level visual features, extracted by an off-the-shelf \acl{CNN}.
Despite we only conducted an offline evaluation of the method, the fixation detection has been designed to follow an online execution paradigm.
 
The method was evaluated on the data collected from intact subjects performing several of the most common grasps in activities of daily living. The acquisition protocol has been designed to simulate the prosthesis usage in a realistic environment. 
To ensure variability, we considered grasps both in a static setting as well as when used to perform a functional task, while we took the limb position effect into account by repeating the movements while seated and standing.
Furthermore, the same objects were associated to multiple grasps to enforce a many-to-many relationship between grasps and objects, 
and multiple objects were placed in the user's field of view to encourage realistic gaze behavior.

Our tests confirmed that the integration of object affordances to the muscular activity of the forearm is indeed useful for grasp classification. The average prediction accuracy went from 80\%, when using only the EMG cue, to 84\%, when integrating EMG and vision. This improvement was considerable, as it involved uniformly all the subjects and all the grasp types. As expected, the contribution of vision was higher at the onset and the offset of the grasp, when the myoelectric cue is affected by motion artifacts. Finally, the analysis of the \acl{MER} suggested that the performances of the multimodal classifier can be further reduced with a majority vote of the predictions at no expense of the prediction delay.


{\small
\bibliographystyle{IEEEtranN} 
\bibliography{icra18}%
}

\end{document}